\pdfoutput=1

\documentclass[11pt]{article}

\usepackage[final]{acl}

\usepackage{times}
\usepackage{latexsym}

\usepackage[T1]{fontenc}

\usepackage[utf8]{inputenc}

\usepackage{microtype}

\usepackage{inconsolata}

\usepackage{graphicx}

\usepackage{amsmath}
\usepackage{amsfonts}
\usepackage{multicol}
\usepackage{multirow}
\usepackage[acronym]{glossaries} 
\glsdisablehyper

\usepackage{tcolorbox}

\usepackage{tikz}
\usepackage{xstring}  %
\usepackage{amssymb}
\usetikzlibrary{shapes,arrows,positioning,calc,decorations.markings}
\tikzset{
  robotbox/.style={
    rectangle, draw=darkgray, fill=lightgray,
    minimum width=7.0cm, minimum height=3.5cm,
    rounded corners=8pt, thick
  },
  samplearrow/.style={
    ->, >=stealth, thick, darkgray,
    shorten >=1pt, shorten <=1pt
  },
  connectarrow/.style={
    ->, >=stealth, line width=0.8mm, darkgray,
    shorten >=1pt, shorten <=1pt
  },
  maskarrow/.style={
    ->, >=stealth, very thick, darkgray,
    shorten >=1pt, shorten <=1pt
  },
  learnarrow/.style={
    thick, darkgray,
    shorten <=2pt, shorten >=2pt,
    postaction={
      decorate,
      decoration={
        markings,
        mark=at position 1 with {\arrow[scale=1]{>}}
      }
    }
  },
  heatnode/.style={
    minimum size=8mm, inner sep=0,
    draw=darkgray, line width=0.5pt,
    font=\small\ttfamily, align=center,
    fill=white
  }
}

\definecolor{validsoft}{RGB}{200,230,200}
\definecolor{invalidsoft}{RGB}{250,200,200}
\definecolor{lightgray}{RGB}{240,240,240}
\definecolor{darkgray}{RGB}{100,100,100}
\definecolor{purple}{RGB}{156,39,176}

\usepackage[capitalize,noabbrev]{cleveref}

\newacronym{llm}{LLM}{Large Language Model}
\newacronym{lm}{LM}{Large Model}
\newacronym{mcts}{MCTS}{Monte-Carlo Tree-Search}
\newacronym{cfg}{CFG}{Context-Free Grammar}
\newacronym{csg}{CSG}{Context-Sensitive Grammar}
\newacronym{cf}{CF}{Context-Free}
\newacronym{cs}{CS}{Context-Sensitive}
\newacronym{asp}{ASP}{Answer Set Programming}
\newacronym{asg}{ASG}{Answer Set Grammar}
\newacronym{mdp}{MDP}{Markov Decision Process}
\newacronym{bon}{BoN}{Best-of-N}
\newacronym{cot}{CoT}{Chain-of-Thought}
\newacronym{tot}{ToT}{Tree-of-Thought}
\newacronym{sgs}{SGS}{Synthetic Grammar Synthesis}
\newacronym{cr}{CR}{Combinatorial Reasoning}

\title{Learning and Enforcing Context-Sensitive Control for LLMs}

\author{
  \textbf{Mohammad Albinhassan\textsuperscript{1}},
  \textbf{Pranava Madhyastha\textsuperscript{2,4}},
  \textbf{Mark Law\textsuperscript{3}},
  \textbf{Alessandra Russo\textsuperscript{1,4}}
\\
\\
  \textsuperscript{1}Imperial College London,
  \textsuperscript{2}City University of London,
  \textsuperscript{3}ILASP Limited, UK \\
  \textsuperscript{4}The Alan Turing Institute
\\
  \texttt{\{m.albinhassan23, a.russo\}@imperial.ac.uk,} \\
  \texttt{pranava.madhyastha@city.ac.uk, mark@ilasp.com}
\\
  \small{
    \textbf{Correspondence:} \href{mailto:m.albinhassan23@imperial.ac.uk}{m.albinhassan23@imperial.ac.uk}
  }
}

\begin{document}
\maketitle

\begin{abstract}

Controlling the output of Large Language Models (LLMs) through context-sensitive constraints has emerged as a promising approach to overcome the limitations of Context-Free Grammars (CFGs) in guaranteeing generation validity. However, such constraints typically require manual specification---a significant barrier demanding specialized expertise. We introduce a framework that automatically learns context-sensitive constraints from LLM interactions through a two-phase process: syntactic exploration to gather diverse outputs for constraint learning, followed by constraint exploitation to enforce these learned rules during generation. Experiments demonstrate that our method enables even small LLMs (1B parameters) to learn and generate with perfect constraint adherence, outperforming larger counterparts and state-of-the-art reasoning models. This work represents the first integration of context-sensitive grammar learning with LLM generation, eliminating manual specification while maintaining generation validity.

\end{abstract}

\section{Introduction}
\begin{figure*}[t]
  \centering
  \resizebox{0.9\textwidth}{!}{
  \begin{tikzpicture}[node distance=1cm, scale=0.9]

    \node[font=\Large] at (0,16)  {\textbf{Phase 1: Syntactic Exploration}};
    \node[font=\Large] at (8.8,16)  {\textbf{Phase 2: Constraint Exploitation}};
    \draw[thick,dotted] (4.25,16.5) -- (4.25,6.5);

    \begin{scope}[yshift=5.5cm]

      \node[robotbox] at (0,7.5) {};
      \node at (3.2,8.8) {\includegraphics[width=1.2cm]{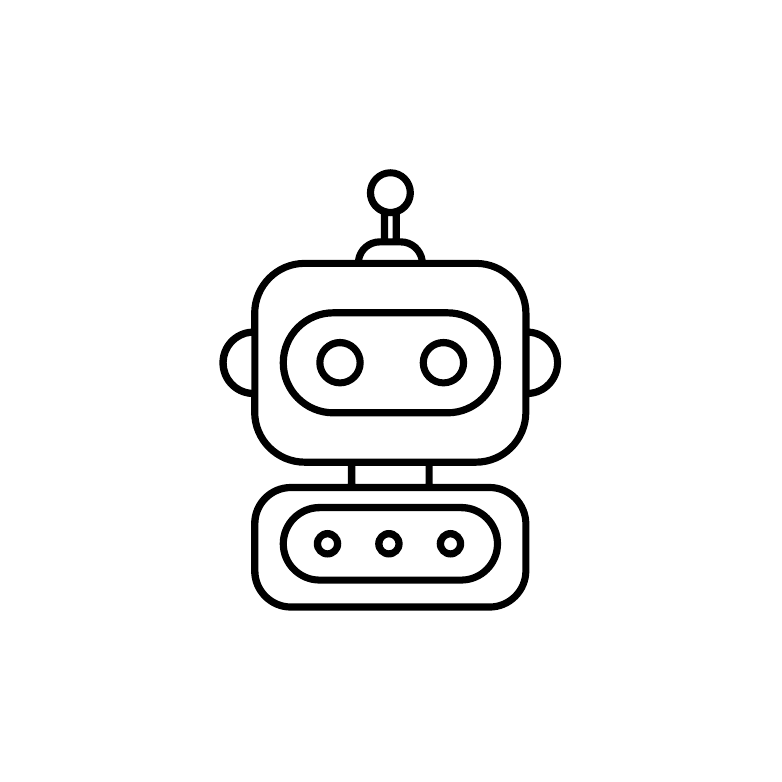}};
      \node[font=\small] at (3.2,8.1) {\bfseries LLM};

      \node[font=\large,anchor=west] at (-3.8,8.2) {$p_\theta$:};
      \node[font=\large,anchor=west] at (-3.8,6.8) {$q_{C_{CFG}}$:};

      \draw[thick,darkgray]
        (-2.4+0.8-0.4,8.2+0.4) rectangle (-2.4+5*0.8+0.4,8.2-0.4);
      \draw[thick,darkgray]
        (-2.4+0.8-0.4,6.8+0.4) rectangle (-2.4+5*0.8+0.4,6.8-0.4);

      \foreach \col/\tok in {1/a,2/cc,3/bc,4/aab,5/cb}{
        \node[heatnode,scale=1.111] at (-2.4+\col*0.889,8.2) {\texttt{\tok}};
      }
      \foreach \validity/\col/\tok in {
        valid/1/a, valid/2/cc,
        valid/3/bc,   valid/4/aab,
        invalid/5/\;cb
      }{
        \IfStrEq{\validity}{valid}{
          \node[heatnode,fill=validsoft,scale=1.111]  at (-2.4+\col*0.889,6.8) {\texttt{\tok}};
        }{
          \node[heatnode,fill=invalidsoft,scale=1.111] at (-2.4+\col*0.889,6.8) {\texttt{\tok}};
        }
      }
      \foreach \c in {1,2,3,4,5}{
        \draw[maskarrow] (-2.4+\c*0.889,7.6) -- (-2.4+\c*0.889,7.4);
      }

      \node[font=\large,anchor=west] at (-3.8,4.5) {$\mathcal V$:};
      \node at (-2.3,4.5) {\small\texttt{abc}   \Large\textcolor{green}{$\checkmark$}};
      \node at (-0.8,4.5) {\small\texttt{aabbcc}\Large\textcolor{green}{$\checkmark$}};
      \node at (1.3,4.5)  {\small\texttt{bcc}   \Large\textcolor{red}{$\times$}};
      \node at (3.3,4.5)  {\small\texttt{aabc}  \Large\textcolor{red}{$\times$}};

      \draw[samplearrow] (-2,5.6)  -- (-2.5,4.8);
      \draw[samplearrow] (-0.7,5.6)-- (-0.7,4.8);
      \draw[samplearrow] (0.7,5.6)  -- (1.0,4.8);
      \draw[samplearrow] (2,5.6)    -- (3.0,4.8);
      \node[font=\large,anchor=west] at (-3.8,5.3) {$g:$};

      \node at (0,2.3)             {\includegraphics[width=1.5cm]{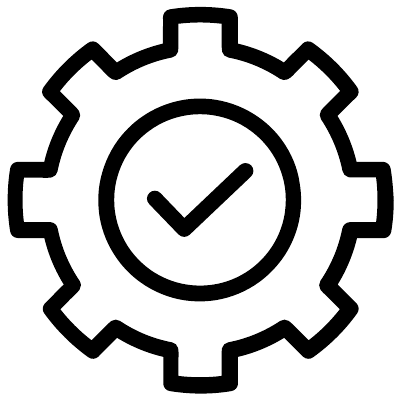}};
      \node[font=\small] at (0,1.3) {\bfseries \acrshort{asg} Learner};
      \node[font=\Large] at (3,2.5) {$\mathbf{\hat G}$};

      \draw[learnarrow] (-2.3,4.2) to[out=-90,in=135] (0,3.3);
      \draw[learnarrow] (-1.0,4.2) to[out=-90,in=140] (0,3.3);
      \draw[learnarrow] (1.3,4.2)  to[out=-90,in=45]  (0,3.3);
      \draw[learnarrow] (3.3,4.2)  to[out=-90,in=45]  (0,3.3);

      \draw[darkgray,line width=0.8pt]
        (-0.15,3.3) -- (0,3.1) -- (0.15,3.3);

      \node[font=\large,anchor=west] at (-3.8,3.2) {$E:$};
      \draw[learnarrow] (0.9,2.5) -- (2.7,2.5);

      \begin{scope}[xshift=8.5cm]
        \node[robotbox] at (0,7.5) {};
        \node at (3.2,8.8) {\includegraphics[width=1.2cm]{images/Robot.pdf}};
        \node[font=\small] at (3.2,8.1) {\bfseries LLM};

        \node[font=\large,anchor=west] at (-3.8,8.2) {$p_\theta$:};
        \node[font=\large,anchor=west] at (-3.8,6.8) {$q_{\hat C_{G_{\mathrm{ASG}}}}$:};

        \draw[thick,darkgray]
          (-2.4+0.8-0.4,8.2+0.4) rectangle (-2.4+5*0.8+0.4,8.2-0.4);
        \draw[thick,darkgray]
          (-2.4+0.8-0.4,6.8+0.4) rectangle (-2.4+5*0.8+0.4,6.8-0.4);

        \foreach \col/\tok in {1/a,2/cc,3/bc,4/aab,5/cb}{
          \node[heatnode,scale=1.111] at (-2.4+\col*0.889,8.2) {\texttt{\tok}};
        }
        \foreach \validity/\col/\tok in {
          valid/1/a, invalid/2/cc,
          invalid/3/bc, valid/4/aab,
          invalid/5/\;cb
        }{
          \IfStrEq{\validity}{valid}{
            \node[heatnode,fill=validsoft,scale=1.111]
              at (-2.4+\col*0.889,6.8) {\texttt{\tok}};
          }{
            \node[heatnode,fill=invalidsoft,scale=1.111]
              at (-2.4+\col*0.889,6.8) {\texttt{\tok}};
          }
        }
        \foreach \c in {1,2,3,4,5}{
          \draw[maskarrow] (-2.4+\c*0.889,7.6) -- (-2.4+\c*0.889,7.4);
        }

        \node at (-2.7,4.5) {\small\texttt{aabbcc}   \Large\textcolor{green}{$\checkmark$}};
        \node at (-1.1,4.5) {\small\texttt{\qquad aaabbcccc}\Large\textcolor{green}{$\checkmark$}};
        \node at (1.3,4.5)  {\small\texttt{abc}      \Large\textcolor{green}{$\checkmark$}};
        \node at (3.3,4.5)  {\small\texttt{aaaabbbbcccc}\Large\textcolor{green}{$\checkmark$}};

        \draw[samplearrow] (-2,5.6)  -- (-2.5,4.8);
        \draw[samplearrow] (-0.7,5.6)-- (-0.7,4.8);
        \draw[samplearrow] (0.7,5.6)  -- (1.0,4.8);
        \draw[samplearrow] (2,5.6)    -- (3.0,4.8);

      \draw[learnarrow] (-2.8,4.2) to[out=-90,in=135] (0,3.3);
      \draw[learnarrow] (-1.0,4.2) to[out=-90,in=140] (0,3.3);
      \draw[learnarrow] (1.1,4.2)  to[out=-90,in=45]  (0,3.3);
      \draw[learnarrow] (3.3,4.2)  to[out=-90,in=45]  (0,3.3);

       \node at (0,2.3)             {\includegraphics[width=1.5cm]{images/gear.pdf}};
      \node[font=\small] at (0,1.3) {\bfseries \acrshort{asg} Learner};
      \node[font=\Huge] at (3.3,2.5) {$\textcolor{green}{\checkmark}$};
      \node[font=\large] at (3.3,1.7) {$\textcolor{black}{\texttt{100\%}}$};
      \draw[darkgray,line width=0.8pt]
        (-0.15,3.3) -- (0,3.1) -- (0.15,3.3);
      \draw[learnarrow] (0.9,2.5) -- (2.7,2.5);
      \end{scope}

    \end{scope}
  \end{tikzpicture}
  }
\caption{Two-phase methodology for learning context-sensitive constraints. \textbf{Phase 1:} An \acrshort{llm} $p_\theta$ samples diverse sequences using generator $g$ from a \acrshort{cfg}-masked distribution ($q_{C_{\text{\acrshort{cfg}}}}$). Tokens such as \texttt{bc} are \acrshort{cfg}-valid but context-sensitively invalid, leading to oracle rejection ($\mathcal{V}$) and red masking in  $\hat{\mathcal{C}}_{\text{\acrshort{asg}}}$. Valid tokens appear green, invalid tokens red. Labeled examples form dataset $E$ for the \acrshort{asg} learner to construct $\hat{G}$. \textbf{Phase 2:} The \acrshort{llm} uses the learned \acrshort{asg}-constrained distribution ($q_{\hat{C}_{G_{\mathrm{ASG}}}}$), disallowing tokens that may lead to violations (red), while valid tokens remain accessible (green), ensuring all outputs satisfy the target grammar (\checkmark). The gear in Phase 2 illustrates all constraints have been learned (100\%), so nothing new is learned (note: this is for visualization purposes only).}\label{fig:method}
\end{figure*}
\acrfullpl{llm} have revolutionized natural language processing, demonstrating unprecedented capabilities across diverse domains \citep{brown2020gpt3, dubey2024llama}. However, ensuring correctness in \acrshort{llm} outputs remains a critical challenge, particularly when outputs must adhere to specific formal constraints. While recent advances in controlled decoding have enabled enforcement of syntactic correctness through \acrfullpl{cfg} \citep[interalia]{geng-etal-2023-grammar,pmlr-v235-beurer-kellner24a, park2024grammaraligned}, ensuring semantic validity requires additional mechanisms.

The fundamental limitation lies in the expressivity gap between \acrshortpl{cfg} and real-world requirements. Many domains demand not only local structural correctness but also relationships between distant elements in a sequence, nested structures, and so on \citep{scholak-etal-2021-picard}. Such constraints can only be expressed by more powerful formalisms like \acrfullpl{csg}. For instance, a \acrshort{cfg} may capture the language $a^ib^jc^k$, where any number of \texttt{a}'s must be followed by any number of \texttt{b}'s and then \texttt{c}'s, but only a \acrshort{csg} can capture dependencies such as equal counts, i.e., $a^nb^nc^n$. Consequently, domain-specific solutions were proposed for tasks like semantic parsing \citep{lei2025spider, poesia2022synchromesh, roy2023benchclamp}, and later, general domain-independent frameworks have been developed \citep{semctrl} to broaden applicability. However, a barrier to adoption exists, as formal specifications for context-sensitive constraints demand expertise that may not be readily available. This contrasts with \acrshortpl{cfg}, which are more widely accessible for many structured generation tasks \citep{wang2023grammar}.

We introduce a framework that automatically learns context-sensitive constraints from \acrshort{llm} outputs. Our approach operates in two phases (\cref{fig:method}): (1) \textit{syntactic exploration}, where we leverage a \acrshort{cfg}-constrained temperature-sampling mechanism to collect diverse syntactically valid outputs, which are then labeled by an oracle and used to learn context-sensitive constraints through a logic-based learner; and (2) \textit{constraint exploitation}, where these learned constraints control \acrshort{llm} generation to guarantee context-sensitive correctness. This represents the first integration of context-sensitive grammar learning with \acrshort{llm} generation.

Our empirical results on synthetic grammar synthesis tasks demonstrate our framework can successfully learn the ground-truth context-sensitive constraints via \acrshort{llm} interactions. As such, our approach induces control in \acrshort{llm} generations and guarantees constraint adherence for even small models (i.e., 1B parameters) --- a capability even state-of-the-art reasoning models (i.e., DeepSeek-R1 \citep{guo2025deepseek}) fail to achieve consistently.

\section{Related Work}

Significant work in controlled decoding has focused on \acrshort{cfg}-based approaches \citep[interalia]{Beurer_Kellner_2023,willard2023efficient}, where \acrshort{llm} generations must conform to the grammar's specification \citep{welleck2024decoding}. These methods address syntactic validity but are unable to enforce context-sensitive constraints critical for many real-world tasks. Semantic parsing via \acrshortpl{llm} aim to capture such constraints; however, they employ domain-specific rules \citep{scholak-etal-2021-picard, roy2023benchclamp, poesia2022synchromesh}. Recent work develops a unifying domain-independent framework for controlling \acrshort{llm} outputs according to \acrshortpl{csg} and semantic constraints via \acrfullpl{asg} \citep{semctrl}, though these constraints remain handcrafted.

\citet{wang2023grammar} propose grammar prompting, where an \acrshort{llm} predicts \acrshortpl{cfg} for specific tasks to control generation. However, the approximated grammar remains context-free and may be incorrect. In contrast, we extend \citet{semctrl} by automatically learning context-sensitive constraints expressed as formal annotations over \acrshortpl{cfg}. These constraints are learned via a state-of-the-art logic-based learner using \acrshort{llm}-generated examples labeled by an oracle. Thus, adapting to new tasks without handcrafting constraints with guaranteed correctness on the learned grammar.

\section{Background}

\textbf{Formal Languages} A formal language $L \subseteq \Sigma^*$ is a set of strings composed of a vocabulary $\Sigma$. $L$ is generated by a grammar $G = \langle N, T, P, S \rangle$ where $N$ are non-terminals, $T = \Sigma$ are terminals, $P$ are production rules, and $S \in N$ is the start symbol. \acrshortpl{cfg} compose of rules of the form $A \rightarrow \alpha$ where $A \in N, \alpha \in (N \cup T)^*$, allowing them to capture syntax. While \acrshortpl{csg} encode rules of the form $\alpha A \beta \rightarrow \alpha \gamma \beta$ where $A \in N$, $\alpha, \beta \in (N \cup T)^*$, $\gamma \in (N \cup T)^+$. Hence, \acrshortpl{csg} can capture context-dependent patterns \citep{linz2022introduction}. As such, while a \acrshort{cfg} captures $L_1 = \{a^i b^j c^k : i,j,k \geq 0\}$, only a \acrshort{csg} can express $L_2 = \{a^n b^n c^n : n \geq 0\}$.

\textbf{Answer Set Grammars} \acrshortpl{asg} \citep{law2019representing} extend production rules of \acrshortpl{cfg} with context-sensitive constraints expressed in a logic-based language called \acrshort{asp} \citep{lifschitz2019answer}. A string $w$ belongs to the language represented by an \acrshort{asg} $G_{\text{ASG}}$, i.e., $w \in L(G_{\text{ASG}})$, if there exists a parse tree derivation whose logic representation (in \acrshort{asp}) is satisfiable --- meaning a set of logical statements, rules, or constraints must all be true simultaneously. For instance, the \acrshort{cfg} component of an \acrshort{asg} captures $L_1$, and the context-sensitive annotations capture $L_2$ by imposing constraints on the number of occurrences of terminal symbols. These annotations have been shown to be learnable from positive and negative examples of a \acrshort{csg} using the logic-based learner ILASP \citep{ilasp}. For example, given $L_1$, a positive example (i.e., \texttt{aabbcc}) and a negative example (i.e., \texttt{aabc}), ILASP learns constraints for equal counts of \texttt{a}'s, \texttt{b}'s, and \texttt{c}'s.

\section{Methodology}

Our approach learns context-sensitive constraints for language model generation through a two-phase process: \textit{syntactic exploration} and \textit{constraint exploitation}. Syntactic exploration works as follows: (1) Starting with a \acrshort{cfg}, we generate
diverse samples from a syntactically constrained \acrshort{llm} via temperature-sampling (we alter temperature to obtain diverse sequences \citep{renze-2024-effect}); (2) We use an oracle to label the samples into positive ($w \in L(G_{\text{CSG}})$) and negative ($w \notin L(G_{\text{CSG}})$) sets; (3) We feed the labeled examples to the \acrshort{asg} learner to learn the context-sensitive annotations over the given \acrshort{cfg} that covers all samples. For constraint exploitation, we follow \citet{semctrl} to constrain the \acrshort{llm}'s generation to conform to the learned context-sensitive constraints.

\subsection{Syntactic Exploration}
\label{sec:syntax-explore}

\textbf{(1) \acrshort{cfg}-Constrained Diverse Sampling.} To learn the context-sensitive constraints of a target grammar $G_{\text{ASG}}$, we require samples that both satisfy and violate these constraints while maintaining syntactic validity (\cref{fig:method}, left). Let $p_\theta$ denote a language model with parameters $\theta$ that defines a distribution over tokens $p_\theta(y_t | x, y_{<t})$ given input $x$ and context $y_{<t}$. We seek to learn the grammar $\hat{G}_{\text{ASG}}$ by collecting a dataset $\mathcal{D}$ containing both positive ($y \in L(G_{\text{ASG}})$) and negative examples ($y \in L(G_{\text{CFG}}) \setminus L(G_{\text{ASG}})$) of the underlying context-sensitive constraints.

Following \citet{semctrl}, we define a constraint function $\mathcal{C} : \mathcal{V}^* \rightarrow 2^{\mathcal{V}}$ that maps any prefix $y_{<t} = (y_1,\ldots,y_{t-1}) \in \mathcal{V}^*$ to the set of valid next tokens according to a grammar $G$:

\begin{equation}
\label{eq:c-abstract}
\begin{split}
\mathcal{C}(y_{<t}) = \{y_t \in \mathcal{V} \mid \;& \exists\, w \in L(G) : (y_{<t} \circ y_t) \\ &\text{ is a prefix of } w\}
\end{split}
\end{equation}

where $\circ$ denotes token concatenation and $\mathcal{V}$ is the vocabulary of the language model's tokenizer. %

We define a temperature-based syntactically constrained sampling generator to construct $\mathcal{D}$ with sufficient diversity to capture various context-sensitive violations. The sampling generator $g$ with parameters $\phi = \{\mathcal{T}, N, C_{\text{CFG}}\}$ is:
\begin{equation}
\begin{split} 
g(y | x; p_\theta, \phi) = &\bigl\{y^{(n,k)} \sim q_{\mathcal{C}_{\text{CFG}}}(\cdot \mid x; p_\theta, \tau_k), \\ &n \in [N], k \in [|\mathcal{T}|]\bigr\}
\end{split}
\end{equation}
where each $y^{(n,k)}$ is a generated sequence, $\mathcal{C}_{\text{CFG}}$ the constraint function for grammar $G_{\text{CFG}}$, $N$ is the number of sequences per temperature value, and $\mathcal{T} = \{\tau_1, \ldots, \tau_T\}$ is the temperature schedule.

Each sequence is sampled as $y \sim q_{\mathcal{C}_{\text{CFG}}}$, where:

\begin{equation}
\label{eq:masked-temp-sample}
\begin{split}
&q_{\mathcal{C}_{\text{CFG}}}(y_t \mid x, y_{<t}; p_{\theta}, \tau) \propto \\ &\text{exp}\left(\frac{s_{\theta}(y_t \mid x, y_{<t})~\mathbb{I}[y_t \in \mathcal{C}_{\text{CFG}}(y_{<t})]}{\tau}\right)
\end{split}
\end{equation}

where $s_\theta$ is the model logit function, $\tau$ is the temperature parameter, and $\mathbb{I}(\cdot)$ is the indicator function. This guarantees that any sampled sequence belongs to $L(G_{\text{CFG}})$.

For a given task with $M$ problem instances $\{x_i\}^M_{i=1}$, applying this generator to all $x_i \in M$ yields a dataset $\mathcal{D} = \{y_{i,j,k} : i \in [M], j \in [N], k \in [T]\}$, where $|\mathcal{D}| = M \cdot N \cdot |\mathcal{T}|$.

\textbf{(2) Oracle Labeling.} We employ a task-specific oracle $V: \Sigma^* \rightarrow \{0,1\}$ to annotate each generated sequence. The oracle is treated as a deterministic ground truth labeler for the constraints, returning $V(y) = 1$ if $y$ satisfies all constraints and 0 otherwise. This transforms our dataset into:

\begin{equation}
E = \{(y_{i,j,k}, V(y_{i,j,k})) : y_{i,j,k} \in \mathcal{D}\}
\end{equation}

The diversity in temperature sampling ensures positive and negative examples are sufficiently populated, providing the \acrshort{asg} learner with comprehensive coverage of the constraint space. %

\textbf{(3) Constraint Learning via \acrshort{asg} Learner.} We segment $E$ into $E^+$ and $E^-$, containing samples conforming to %
and violating %
the constraints, respectively, as given by the oracle. We feed as input to the \acrshort{asg} learner $G_{\text{CFG}}$, $E^+$, and $E^-$. Consequently, $\hat{G}_{\text{ASG}}$ is constructed by learning the \acrshort{asp} annotations over $G_{\text{CFG}}$ such that $\hat{G}_{\text{ASG}}$ covers all samples in $E$ 
 (see \cref{app:asg} for formal details).

\subsection{Constraint Exploitation}

With the learned \acrshort{asg} $\hat{G}_{\text{ASG}}$, we transition from syntactic exploration to constraint exploitation (\cref{fig:method}, right). Following \citet{semctrl}, we sample sequences $y \sim q_{\hat{\mathcal{C}}_{\text{ASG}}}$ encoding the constraint function $\hat{\mathcal{C}}_{\text{ASG}}$ for the learned grammar $\hat{G}_{\text{ASG}}$. This is similar to \cref{eq:masked-temp-sample} without temperature variations. At this point, the model has no further access to the oracle, relying entirely on the learned constraints to ensure context-sensitive validity.

\section{Experiments}

\subsection{Task Definition}

We evaluate our approach on two synthetic grammar synthesis tasks, where the \acrshort{llm} must generate strings from a target context-sensitive language. Following \citet{semctrl}, we adopt $L_1 = \{a^nb^nc^n \mid n \geq 1\}$ and craft $L_2 = \{a^nb^nc^m \mid n,m \geq 1\}$. Each problem instance $x_i \in M$ prompts the \acrshort{llm} to generate strings with various values of $n$ and $m$, producing diverse examples that capture both valid and invalid patterns with respect to the context-sensitive constraints for the \acrshort{asg} learner.

\subsection{Experimental Setup}

\textbf{Models.} We evaluate closed- and open-source models across various sizes: GPT-4.1, o1, o3-mini, o4-mini, and DeepSeek-R1 through their respective APIs, and Llama models (3.2 1B, 3.2 3B, 3.1 8B, and 3.1 70B) which we run locally (see \cref{app:compute} for GPU cluster details). All models are prompted identically using few-shot examples.

\textbf{\acrshort{asg} Learning Configuration.} We sample 10 generations at each temperature value $\tau \in \{0, 0.1, ..., 1.0\}$ for the syntactic exploration phase to construct a diverse dataset $\mathcal{D}$. The oracle $V(y)$ is implemented as a Python program to check constraint validity, i.e., checks the counts of \texttt{a}'s, \texttt{b}'s, and \texttt{c}'s and their respective ordering. The \acrshort{asg} learner constructs $\hat{G}_{\text{ASG}}$ by learning the \acrshort{asp} annotations over $G_{\text{CFG}}$ from these examples segmented into $E^+$ and $E^-$.

\textbf{Unconstrained and Constraint Exploitation Sampling Mechanisms.} For API-based models, we use their standard generation settings. For Llama models, we employ three sampling approaches: (1) unconstrained rejection sampling, where we generate 50 samples and select a generation based on the oracle's feedback; and constrained generation, where we apply (2) the learned \acrshort{asg} and (3) a handcrafted \acrshort{asg} for comparison with \citet{semctrl}.

\textbf{Evaluation Metrics.} We evaluate methods using context-sensitive validity accuracy, defined as the percentage of generated sequences that belong to the ground-truth grammar $G_{\text{ASG}}$.

\subsection{Results and Analysis}
\label{sec:results}

\begin{table}
  \centering
  \begin{tabular}{lrrr}
  
    \hline
    \multirow{2}{*}{\textbf{Model}} & \multirow{2}{*}{$\mathbf{G}$} & \multicolumn{2}{c}{\textbf{Accuracy}} \\
    \cline{3-4}
    & & $a^nb^nc^n$ & $a^nb^nc^m$ \\
    \hline
    GPT 4.1 & - & 63.3\% & 76.7\% \\
    o1 & - & 86.7\% & 96.7\% \\
    o3 mini & - & 63.3\% & 86.7\% \\
    o4 mini & - & 90.0\% & 93.3\% \\
    DeepSeek-R1 & - & 80.0\% & 86.7\% \\
    \hline
    Llama 1B & - & 20.0\% & 6.7\% \\
    \textbf{Llama 1B} & $\mathbf{G_{\textbf{ASG}}}$ & \textbf{100.0\%} & \textbf{100.0\%} \\
    \textbf{Llama 1B} & $\mathbf{\hat{G}_{\textbf{ASG}}}$ & \textbf{100.0\%} & \textbf{100.0\%} \\
    Llama 70B & - & 76.7\% & 53.3\% \\
    \textbf{Llama 70B} & $\mathbf{G_{\textbf{ASG}}}$ &\textbf{ 100.0\%} & \textbf{100.0\%} \\
    \textbf{Llama 70B} & $\mathbf{\hat{G}_{\textbf{ASG}}}$ &\textbf{ 100.0\%} & \textbf{100.0\%} \\
    \hline
  \end{tabular}
  \caption{Accuracy results for $a^nb^nc^n$ and $a^nb^nc^m$ with different \acrshortpl{llm} (Model) and grammar constraints ($\mathrm{G}$).}
  \label{tab:results}
\end{table}

Table \ref{tab:results} summarizes our findings across models and constraints (see \cref{app:results} for results on 3B and 8B). We analyze two key aspects: the effectiveness of our \acrshort{asg} learning approach, and the impact of learned constraints on accuracy.

\textbf{Ground-Truth \acrshortpl{asg} are Learned.} Table \ref{tab:results} showcases that constraining \acrshort{llm} $p_\theta$ with the ground-truth grammar ($G_{\text{ASG}}$) and the learned grammar ($\hat{G}_{\text{ASG}}$) both provide 100\% accuracy and conform to all constraints. Whilst it could be the case that our sampling mechanism with the \acrshort{asg} learner only learned a subset of constraints sufficient for the \acrshort{llm} not to make any errors, i.e., the \acrshort{llm} already captures some of these via the prompt, manual inspection confirmed $\hat{G}_{\text{ASG}}$ is identical to $G_{\text{ASG}}$. The reasons behind this are twofold: (1) our syntax-constrained temperature-based sampling approach effectively covers the space of context-sensitive constraints sufficiently, i.e., the necessary positive and negative examples; (2) the \acrshort{asg} learner based on ILASP guarantees that all examples will be covered, and if a solution exists, it will be found (see \citet{law2015proof} for soundness and completeness proofs).

\textbf{Guaranteed Correctness via Constraints.} When applying the learned \acrshort{asg} constraints during generation, all models---even the smallest 1B-parameter model---achieve 100\% accuracy on both context-sensitive tasks. In contrast, unconstrained generation with larger and closed-source models fails to provide such guarantees, with Llama 70B achieving only 76.7\% and 53.3\% accuracy, and GPT-4.1 obtaining 63.3\% and 76.7\% on $L_1$ and $L_2$, respectively. Although increasing the scale of model parameters improves performance (e.g., Llama 1B's 20.0\% and 6.7\% vs. Llama 70B), unconstrained models still lack reliability and robustness in generation. %

Despite employing significantly more computational resources through extended reasoning steps \citep{valmeekam2025a, guo2025deepseek, semctrl}, state-of-the-art reasoning models (i.e., o1, DeepSeek-R1, etc.) still produce invalid sequences. Consider o4-mini, the best performing unconstrained model, still only achieves 90.0\% and 93.3\% on $L_1$ and $L_2$, respectively. These results demonstrate that our neuro-symbolic constraint learning approach provides correctness guarantees that cannot currently be achieved through scale or inference time multi-step reasoning alone. Most notably, a 1B-parameter model eliminates the need for handcrafted constraints by learning and enforcing the ground-truth constraints, consistently outperforming all unconstrained models. This emphasizes the complementary strengths of neural language generation and symbolic constraint enforcement.

\section{Conclusion and Future Work}
\label{sec:conclusion}

We presented a novel framework for automating the learning of context-sensitive constraints for controlled \acrshort{llm} generation. The synergistic combination of syntactic exploration and constraint exploitation eliminates the need for manual constraint specification while maintaining correctness guarantees. Our empirical results demonstrate that this method enables small \acrshortpl{llm} to learn and generate with perfect constraint adherence, outperforming larger and specialized reasoning models.

We plan to extend our work to real-world settings where constraints represent semantic relationships with intrinsic meaning (i.e., semantic parsing, agent planning). We further aim to explore active learning settings using \acrshort{asg}'s sample-efficient one-shot learning ability. Thus, enabling continuous constraint refinement in lifelong learning tasks where a complete \acrshort{asg} may not be initially captured. %

\section*{Limitations}

Our approach demonstrates promising results, yet several limitations warrant consideration. First, the syntactic exploration phase lacks formal convergence guarantees. While temperature-based sampling empirically captured sufficient constraint violations in our synthetic domains, we cannot guarantee comprehensive coverage of larger constraint spaces. Establishing theoretical connections between sampling strategies, sample efficiency, and constraint space coverage remains an open challenge.

Second, our framework currently addresses only hard constraints where outputs are strictly valid or invalid. Many real-world NLP tasks, such as machine translation or question answering, involve soft constraints where outputs exist on a spectrum of acceptability. This binary classification approach limits applicability to domains requiring nuanced evaluation of correctness.

Third, our method assumes the underlying language model has been trained on data containing the relevant terminals and has developed statistical priors aligned with the target formal languages. For domains with limited representation in the training corpus, the generated samples may be insufficient to capture the full spectrum of context-sensitive constraints. We acknowledge these limitations and aim to address them in our future work (\cref{sec:conclusion}).

\section{Acknowledgements}

We thank Microsoft Research - Accelerating Foundation Models Research program for the provision of Azure resources to run some of the \acrshortpl{llm} used in the experiments in this paper. This research was partially sponsored by DEVCOM Army Research Lab under W911NF2220243, EPSRC project EP/Y037421/1, and The Alan Turing Institute's project on Robust Inference with PASP scaffolds for \acrshortpl{llm}.

\bibliography{anthology,custom}

\appendix

\section{ASG Example and Learning Details}
\label{app:asg}

\subsection{ASG Example}

\begin{figure}[ht]
\centering
\begin{tcolorbox}[
  colback=gray!7,
  colframe=gray!30,
  boxrule=0.6pt,
  width=0.92\linewidth,
  left=6pt,
  right=6pt,
  top=6pt,
  bottom=6pt]
\begin{minipage}{0.95\linewidth}
\noindent\texttt{start} $\longrightarrow$ \texttt{as}~\texttt{bs}~\texttt{cs} \textcolor{darkgray}{\texttt{\textbf{\{}}}\\
\hspace*{2em} \texttt{\textbf{:- size(X)@1, not size(X)@2.}}\\
\hspace*{2em} \texttt{\textbf{:- size(X)@1, not size(X)@3.}}\\
\texttt{\textbf{\}}}\\[0.8em]

\noindent\texttt{as} $\longrightarrow$ \texttt{"a"}~\texttt{as} \textcolor{darkgray}{\texttt{\textbf{\{}}}\\
\hspace*{2em} \texttt{\textbf{size(X+1) :- size(X)@2.}}\\
\texttt{\textbf{\}}} $\;\;|\;\;$ \textcolor{darkgray}{\texttt{\textbf{\{}}}\\
\hspace*{2em} \texttt{\textbf{size(0).}}\\
\texttt{\textbf{\}}}\\[0.8em]

\noindent\texttt{bs} $\longrightarrow$ \texttt{"b"}~\texttt{bs} \textcolor{darkgray}{\texttt{\textbf{\{}}}\\
\hspace*{2em} \texttt{\textbf{size(X+1) :- size(X)@2.}}\\
\texttt{\textbf{\}}} $\;\;|\;\;$ \textcolor{darkgray}{\texttt{\textbf{\{}}}\\
\hspace*{2em} \texttt{\textbf{size(0).}}\\
\texttt{\textbf{\}}}\\[0.8em]

\noindent\texttt{cs} $\longrightarrow$ \texttt{"c"}~\texttt{cs} \textcolor{darkgray}{\texttt{\textbf{\{}}}\\
\hspace*{2em} \texttt{\textbf{size(X+1) :- size(X)@2.}}\\
\texttt{\textbf{\}}} $\;\;|\;\;$ \textcolor{darkgray}{\texttt{\textbf{\{}}}\\
\hspace*{2em} \texttt{\textbf{size(0).}}\\
\texttt{\textbf{\}}}
\end{minipage}
\end{tcolorbox}
\caption{The learned \acrshort{asg} for $a^nb^nc^n$ using our approach. This grammar utilizes \acrshort{asp} constraints (in bold and surrounded by $\{\}$) to enforce the context-sensitive condition that all three symbol sequences maintain equal length.}
\label{fig:asg-example-anbncn}
\end{figure}

Figure~\ref{fig:asg-example-anbncn} illustrates the \acrshort{asg} learned via the \acrshort{asg} learner based on ILASP for the language $L= \{a^nb^nc^n : n \geq 1\}$. The \acrshort{asg} consists of two key aspects:

\begin{enumerate}
    \item A \acrshort{cfg} expressed in Extended Backus–Naur form, i.e., \texttt{as $\rightarrow$ ``a" as}. Here, the non-terminals are \texttt{as}, \texttt{bs}, and \texttt{cs}, the terminals are \texttt{a}, \texttt{b}, and \texttt{c}, the start symbol is \texttt{start}, and $\rightarrow$ denotes the production rules (i.e., the non-terminal on the left-hand side of the arrow can be replaced by the terminal on the right-hand side of the arrow).
    \item Context-sensitive constraints annotating the production rules expressed in \acrshort{asp} code (for further details on \acrshort{asp}, please see \citet{lifschitz2019answer}). The constraints are encoded via curly braces $\{\dots\}$ in the \acrshort{asg} and illustrated in bold text. The first rule's constraints enforce that all three non-terminals must generate sequences of equal length by requiring \texttt{size(X)} to be consistent across all child positions. Terminal productions implement a counting mechanism where each recursive rule increments the size counter by one, while base cases initialize \texttt{size(0)}. The \texttt{@} symbol refers to specific child positions in productions and parse trees, enabling position-dependent constraint checking. For example, \texttt{size(X)@1} refers to the count accumulated in the first child of the parse tree.
\end{enumerate}

\subsection{Constraint Learning via ASG Learner and ILASP.} 

\cref{sec:syntax-explore} provides an intuitive description of how the \acrshort{asg} learner, based on ILASP, learns the context-sensitive constraints. Following \cite{law2019representing}, we now formally define an \acrshort{asg} learning task as $T = \langle G_{\text{CFG}}, S_M, \langle E^+, E^- \rangle \rangle$. Here, $G_{\text{CFG}}$ serves as the base \acrshort{cfg} grammar,  $S_M$ is the search space of possible \acrshort{asp} annotations on production rules to construct $G_{\text{ASG}}$, and $E^+, E^-$ are positive and negative examples, respectively.

Given these inputs, ILASP learns a minimal hypothesis $H \subseteq S_M$ containing \acrshort{asp} annotations over $G_{\text{CFG}}$ such that:
\begin{align}
\forall y \in E^+: y \in L(G_{\text{CFG}} : H) \\
\forall y \in E^-: y \notin L(G_{\text{CFG}} : H)
\end{align}

where $G_{\text{CFG}} : H$ denotes the \acrshort{asg} ($G_{\text{ASG}}$) constructed by extending $G_{\text{CFG}}$ with annotations from $H$. The learned constraints in $H$ encode context-sensitive rules (e.g., enforcing $\text{count}(a) = \text{count}(b) = \text{count}(c)$ for $a^nb^nc^n$ as in \cref{fig:asg-example-anbncn}). Given ILASP searches for a solution covering all examples, we remove duplicate samples when we feed $E^+$ and $E^-$ to the \acrshort{asg} learner.

\section{Further Results}
\label{app:results}

\begin{table}
  \centering
  \begin{tabular}{lrrr}
    \hline
    \multirow{2}{*}{\textbf{Model}} & \multirow{2}{*}{$\mathbf{G}$} & \multicolumn{2}{c}{\textbf{Accuracy}} \\
    \cline{3-4}
    & & $a^nb^nc^n$ & $a^nb^nc^m$ \\
    \hline
    GPT 4.1 & - & 63.3\% & 76.7\% \\
    o1 & - & 86.7\% & 96.7\% \\
    o3 mini & - & 63.3\% & 86.7\% \\
    o4 mini & - & 90.0\% & 93.3\% \\
    DeepSeek-R1 & - & 80.0\% & 86.7\% \\
    \hline
    Llama 1B & - & 20.0\% & 6.7\% \\
    \textbf{Llama 1B} & $\mathbf{G_{\textbf{ASG}}}$ & \textbf{100.0\%} & \textbf{100.0\%} \\
    \textbf{Llama 1B} & $\mathbf{\hat{G}_{\textbf{ASG}}}$ & \textbf{100.0\%} & \textbf{100.0\%} \\
    Llama 3B & - & 20.0\% & 23.3\% \\
    \textbf{Llama 3B} & $\mathbf{G_{\textbf{ASG}}}$ & \textbf{100.0\%} & \textbf{100.0\%} \\
    \textbf{Llama 3B} & $\mathbf{\hat{G}_{\textbf{ASG}}}$ & \textbf{100.0\%} & \textbf{100.0\%} \\
    Llama 8B & - & 46.7\% & 10.0\% \\
    \textbf{Llama 8B} & $\mathbf{G_{\textbf{ASG}}}$ & \textbf{100.0\%} & \textbf{100.0\%} \\
    \textbf{Llama 8B} & $\mathbf{\hat{G}_{\textbf{ASG}}}$ & \textbf{100.0\%} & \textbf{100.0\%} \\
    Llama 70B & - & 76.7\% & 53.3\% \\
    \textbf{Llama 70B} & $\mathbf{G_{\textbf{ASG}}}$ &\textbf{ 100.0\%} & \textbf{100.0\%} \\
    \textbf{Llama 70B} & $\mathbf{\hat{G}_{\textbf{ASG}}}$ &\textbf{ 100.0\%} & \textbf{100.0\%} \\
    \hline
  \end{tabular}
  \caption{Accuracy results for $a^nb^nc^n$ and $a^nb^nc^m$ with different \acrshortpl{llm} (Model), including Llama 3.2 3B and 3.1 8B, and grammar constraints ($\mathrm{G}$).}
  \label{tab:more-results}
\end{table}

\cref{sec:results} showcased context-sensitive accuracy results with respect to various \acrshortpl{llm} and grammar constraints. Here, \cref{tab:more-results} presents results with Llama 3.2 3B and Llama 3.1 8B, which we omitted from the main text due to space requirements. Similar conclusions can be drawn as before. Hence, we omit any further discussions.

\section{GPU Specification}
\label{app:compute}
Our experiments were conducted using a GPU cluster with nodes containing 2× Intel Xeon Platinum 8358 CPUs (2.60GHz, 32 cores each) and NVIDIA L40S GPUs (48GB GDDR6), where we utilized up to 4 GPUs with 96GB RAM.

\onecolumn
\section{Prompt Example}
\label{app:prompts}

\begin{figure}[ht]
\begin{tcolorbox}[colback=gray!7,colframe=gray!25,boxrule=0.6pt,left=8pt,right=8pt,top=8pt,bottom=8pt]
\small
\textbf{System Instruction:}\\ \\
You are an expert in formal languages, specifically, Context-Free and Context-Sensitive Grammars. You can read and understand grammars, and given a grammar specification, you can generate words that consistently conform to the grammar, its language, and rules without a single mistake. For each message, generate a word (a sequence of characters belonging to the language) that conforms to the grammar specification a\textsuperscript{n}b\textsuperscript{n}c\textsuperscript{n}. This grammar represents the language of strings consisting of n number of a's, followed by n number of b's, and finally followed by n number of c's, where all n's are equal (i.e., the number of a's, b's, and c's are all equal) and in the specified order. Each message will specify a max n value, meaning, the individual number of a's, b's, and c's cannot exceed that amount, and you must aim to maximise n (length of words) up to the specified max, thereby, prefering longer words of the grammar's language.\\ \\
Only respond with a single word that conforms to the grammar, do not generate any additional text beyond the correct word with respect to the grammar.
\vspace{0.6em}

\textbf{Example Interactions:}\\ \\
\textit{\textbf{User:}} Generate a valid word/string of the grammar a\textsuperscript{n}b\textsuperscript{n}c\textsuperscript{n}, where you should prefer larger numbers of n (i.e., longer sequences) and the max n value is 3. 

\textit{\textbf{Assistant:}} \texttt{aaabbbccc} \\

\textit{\textbf{User:}} Generate a valid word/string of the grammar a\textsuperscript{n}b\textsuperscript{n}c\textsuperscript{n}, where you should prefer larger numbers of n (i.e., longer sequences) and the max n value is 10.

\textit{\textbf{Assistant:}} \texttt{aaaaaaaaaabbbbbbbbbbcccccccccc}
\end{tcolorbox}
\caption{Prompt template for the a\textsuperscript{n}b\textsuperscript{n}c\textsuperscript{n} language generation task. The system instruction defines the formal language requirements, followed by example interactions demonstrating expected inputs and outputs.}
\label{fig:anbncn-prompt}
\end{figure}

\cref{fig:anbncn-prompt} illustrates the prompt used for the $a^nb^nc^n$ task, with a similar style for our $a^nb^nc^m$ task. Akin to \citet{semctrl}, we adopt a standard few-shot prompting strategy, where we provide a description of the task, syntax, and constraints in natural language and formal language notation.

\end{document}